\documentclass[letterpaper, 10 pt, conference]{ieeeconf}

\IEEEoverridecommandlockouts 

\let\labelindent\relax
\usepackage{enumitem}

\usepackage[english]{babel}
\usepackage[utf8x]{inputenc}
\usepackage{amsmath}

\usepackage{graphicx}
\usepackage[colorinlistoftodos]{todonotes}
\usepackage{cite}
\usepackage{algorithm}
\usepackage{algorithmic}
\usepackage[hidelinks]{hyperref}
\usepackage{amsfonts}
\usepackage{pgfplots}
\pgfplotsset{compat=newest}
\usepgfplotslibrary{groupplots}
\usepackage{newunicodechar}
\usepackage{booktabs}
\usepackage{float} 
\usepackage{multirow}
\usepackage{graphicx}

\usepackage{xcolor}
\definecolor{palGreen1}{RGB}{68,170,139}   
\definecolor{palTeal}{RGB}{77,144,143}     
\definecolor{palBlue}{RGB}{86,116,144}     
\definecolor{palDark}{RGB}{39,70,82}       
\definecolor{palOrange}{RGB}{247,149,30}   
\definecolor{palGold}{RGB}{233,195,106}    

\usepackage{tikz}
\usepackage{pgfplots}
\pgfplotsset{compat=1.18}

\title{GeoAware-VLA: Implicit Geometry Aware Vision-Language-Action Model}

\newif\ifanonym
\anonymfalse 

\usepackage{censor}
\usepackage[hidelinks]{hyperref}
\ifanonym
  \hypersetup{pdfauthor={},pdftitle={},pdfsubject={},pdfkeywords={}}
\fi

\author{%
\ifanonym
  \censor{Anonymous Author(s)}%
  \thanks{\censor{Affiliation(s) and correspondence redacted for double-anonymous review.}}%
\else
  Ali Abouzeid$^{1}$, 
  Malak Mansour$^{1}$,
  Qinbo Sun$^{1}$, 
  Zezhou Sun$^{1}$, 
  Dezhen Song$^{1}$%
  \thanks{$^{1}$Authors are with the Department of Robotics,
        Mohamed bin Zayed University of Artificial Intelligence, Masdar City, Abu Dhabi, UAE. Correspondence to: \tt\small ali.abouzeid@mbzuai.ac.ae}%
\fi
}

\begin{document}
\maketitle

\begin{abstract}
Vision-Language-Action (VLA) models often fail to generalize to unseen camera viewpoints, a limitation stemming from their difficulty in inferring robust 3D geometry from 2D images. We introduce GeoAware-VLA, a simple yet effective approach that enhances viewpoint invariance by integrating strong geometric priors into the vision backbone. Instead of training a visual encoder or relying on explicit 3D data, we leverage a frozen, pretrained geometric vision model as a feature extractor. A lightweight, trainable projection layer then adapts these geometrically-rich features for the policy decoder, relieving it of the burden of learning 3D consistency from scratch. Through extensive evaluations on the LIBERO and CALVIN benchmarks, we show that GeoAware-VLA preserves and even improves in-distribution performance while achieving substantial gains in zero-shot generalization to unseen camera poses, improving unseen-view success rates by an average of 35 percentage points on LIBERO and over 11 percentage points on CALVIN compared to their respective baselines. Crucially, these gains transfer to the physical world, where our model shows significant improvement on a real robotic platform. Our approach proves effective across both continuous and discrete action spaces, highlighting that robust geometric grounding is a key ingredient for building more generalizable robotic agents. 
\end{abstract}


\section{Introduction}

The development of general-purpose agents capable of performing diverse manipulation tasks in unstructured environments is a central goal in robotics.
Vision-Language-Action (VLA) model is a popular paradigm, which learns to map visual observations and natural language instructions directly to robot actions. 
Most methods achieve strong performance within their training domains~\cite{haldar2024baku, kim2024openvla, zitkovich2023rt}, but often fail to generalize beyond them, struggling even with minor changes in camera viewpoints~\cite{simpler, xie2024decomposing}.
This limitation arises from the difficulty of inferring a consistent 3D world model from 2D visual inputs, a prerequisite for reliable manipulation and spatial awareness.

Prior work has explored two main strategies to address this challenge. One approach is to incorporate explicit 3D representations, such as point clouds, into the policy's observation space \cite{wilcox2025adapt3r, yang2025fp3, ze20243d}. While effective, these methods often require depth sensors and introduce significant computational overhead for constructing and processing explicit 3D structures. An alternative lies in implicitly encouraging the model to learn geometrically consistent features without reconstructing the scene in 3D. This is often achieved by leveraging multi-view data or data augmentation during training to produce a view-invariant policy \cite{pang2025learning, tian2024view}.

\begin{figure}[t]
    \centering   \includegraphics[width=\linewidth]{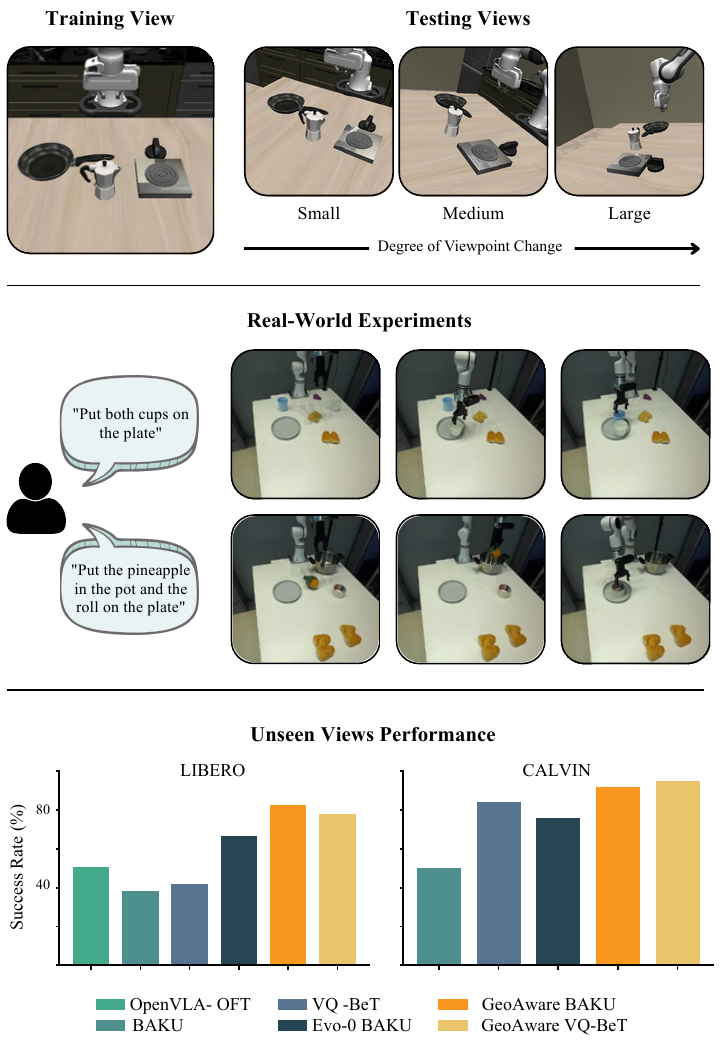}
    \caption{\textbf{(Top)} Illustration of the training and testing viewpoints in the LIBERO dataset. GeoAware-VLA demonstrates zero-shot generalization across unseen viewpoints. \textbf{(Mid)} Key intermediate steps during successful real-world deployment. \textbf{(Bottom)} Average success rates on unseen viewpoints across the LIBERO and CALVIN benchmarks. GeoAware-VLA consistently outperforms all baselines on both benchmarks.}
    \label{fig:main}
\end{figure}

However, these implicit methods also have their own drawbacks. Disentanglement techniques often rely on curated multi-view datasets for effective training, while augmentation-based methods are fundamentally limited by the computational cost of producing unseen views and the distribution of views from which they are sampled. These challenges can be addressed by leveraging a powerful, pretrained geometric foundation model that has already distilled 3D understanding from vast datasets. For instance, the Visual Geometry Grounded Transformer (VGGT) \cite{wang2025vggt} is explicitly optimized to distill a rich 3D representation of the world, having been trained to infer key geometric aspects such as camera parameters, multi-view depth, dense point clouds, and point tracking.

Built on these insights, we propose GeoAware-VLA, a simple yet highly effective modification to a standard VLA architecture that significantly enhances robustness to unseen camera viewpoints. Our core hypothesis is that a policy's ability to generalize across views is fundamentally tied to the geometric precision of its visual encoder. We therefore replace the standard image encoder with the powerful VGGT backbone, using it as a frozen feature extractor. By building upon features that are already inherently consistent across different perspectives, our policy is relieved of the burden of learning 3D geometry from scratch. We need only to append a lightweight, trainable projection layer to map the pretrained features into the latent space of the BAKU policy decoder \cite{haldar2024baku}.

As our experiments demonstrate across both the LIBERO and CALVIN benchmarks, the integration of a powerful geometric prior results in substantial improvements in zero-shot generalization to unseen camera viewpoints, consistently outperforming both standard VLA baselines and recent geometric integration approaches such as Evo-0 \cite{evo0}, as shown in Fig.~\ref{fig:main}.

In summary, our contributions are as follows:

\begin{itemize}
    \item We propose GeoAware-VLA, a simple and effective method for integrating geometric foundation models into VLA architectures via a frozen backbone and lightweight trainable projection layer.
    \item We show that GeoAware-VLA achieves 35\% and 11\% improvements in zero-shot generalization to unseen viewpoints on LIBERO and CALVIN, respectively, and that these gains transfer to a real robotic platform.
    \item We show that our approach is agnostic to the choice of action decoder, consistently improving generalization across both continuous and discrete action spaces.
\end{itemize}

\begin{figure*}[t!]
    \centering    
    \includegraphics[width=0.85\textwidth]{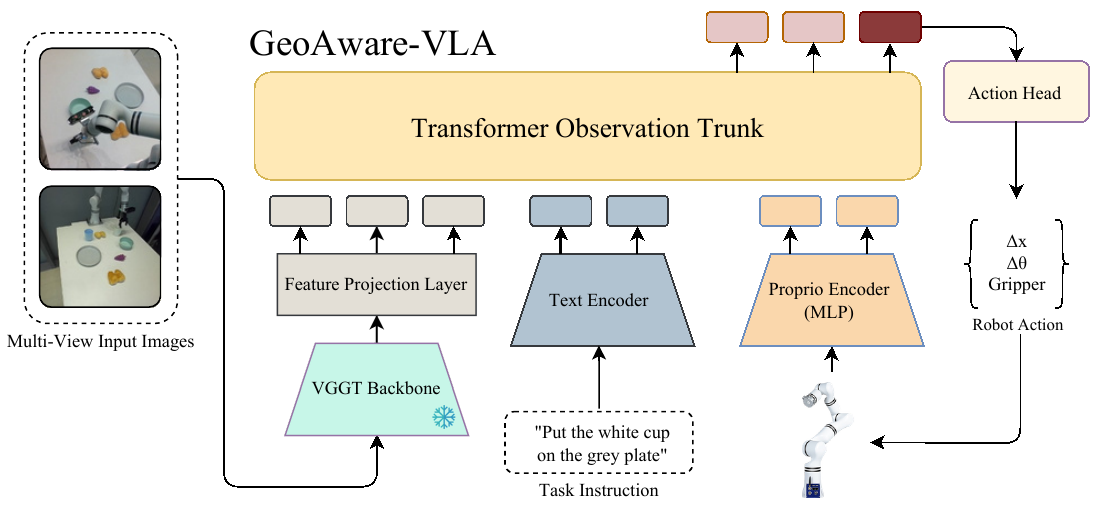}
    \caption{
    The diagram of GeoAware-VLA. It inputs multi-view images, uses VGGT to extract view robust features to form a viewpoint robust policy.}

    \label{fig:model_arch}
\end{figure*}

\section{Related Work}

\subsection{Imitation Learning}

Modern robotic learning is rooted in imitation learning (IL), where early methods such as behavioral cloning (BC) \cite{pomerleau1988alvinn} struggled with compounding errors due to distribution shift. This motivated the development of more robust visuomotor policies using generative methods such as diffusion policy \cite{chi2023diffusion} and energy-based models such as Implicit Behavioral Cloning (IBC) \cite{florence2022implicit}, which learn complex action distributions to enhance control. Building on this robustness, the field's focus shifted towards creating large-scale, generalist agents. Influential models like DeepMind's Gato \cite{reedgeneralist} and Octo \cite{ghosh2024octo} demonstrated that a single, large transformer pre-trained on diverse, multi-robot datasets could establish a powerful and adaptable motor control foundation. This trajectory culminated in the current paradigm of VLA models, which unify perception, language, and control by leveraging knowledge from web-scale models. This includes a series of influential works such as RT-1 \cite{RT-1} and its successors \cite{RT-2, RT-X}, alongside powerful open-source models like OpenVLA \cite{kim2024openvla, kim2025fine} and  $\pi$ -\cite{black2410pi0, intelligence2504pi05}, which have collectively pushed the boundaries of generalization and performance. Our work builds on the BAKU architecture \cite{haldar2024baku}, a lightweight VLA that stands out for its efficient design choices, including multi-sensory observations and action chunking.

\subsection{Vision Encoders in Robotic Learning} 

The vision backbone, which processes raw pixel inputs, is a critical component of VLA models. Foundational approaches often used semantic encoders like ResNet \cite{he2016deep}, and later enhanced with FiLM layers to incorporate conditioning from language or other inputs \cite{perez2018film}.

A more recent paradigm involves models like SigLIP \cite{zhai2023sigmoid} and its successor SigLIP 2 \cite{tschannen2025siglip}, which are pre-trained on vast web-scale image-text pairs. This training method creates representations that are naturally aligned with language, making them highly effective for VLA tasks. However, the primary focus of these backbones remains semantic - identifying what objects are, rather than their precise 3D geometry.

Alongside the development of semantic encoders for policy learning, a distinct class of powerful geometric vision models has emerged, such as DUSt3R \cite{wang2024dust3r} and VGGT \cite{wang2025vggt}. While highly effective for 3D reconstruction, their integration into robot learning is nascent. Most similar to our work, Evo-0 \cite{evo0} explores using VGGT as an auxiliary spatial encoder. However, their architecture retains a standard 2D vision backbone alongside VGGT, using cross-attention to fuse spatial tokens extracted exclusively from VGGT's final layer with the 2D visual features. In contrast, GeoAware-VLA completely replaces the standard image encoder with VGGT as the sole vision backbone. Additionally, rather than relying on a single output layer, we employ a multi-scale projection across intermediate VGGT layers, allowing the policy to leverage both fine-grained geometric details and higher-level representations.

\subsection{View-point robust robotic policies}

To achieve more viewpoint-robust policies, various methods have been explored. One approach involves augmenting training data with images rendered from simulators \cite{pang2025learning, seo2023multi}. While effective, this introduces the challenge of sim-to-real transfer. Another promising way is utilizing novel view synthesis \cite{tian2024view, yang2025mobi}. Unlike simulator-based methods, these models can operate directly on real-world data, circumventing the sim-to-real gap. However, these models may still perform poorly when the new viewpoint is drastically different from the original view \cite{tian2024view}.

Other works, like Act3D \cite{gervet2023act3d} and 3D Diffuser Actor \cite{ke20243ddiffuser}, utilize explicit 3D representations such as point clouds or voxel grids \cite{shridhar2023perceiver, zhulearning, ze20243d, wilcox2025adapt3r}. These methods are robust to viewpoint changes but often require reliable depth data and camera calibration, which can be scarce in large-scale datasets like \cite{RT-X, libero, walke2023bridgedata}. In contrast to these methods, our approach integrates strong geometric priors, which relieves the policy from the burden of learning 3D consistency from scratch and avoids the need for explicit depth or calibration information.

\section{Problem Setup}

We consider a standard setup for multi-task imitation learning where the goal is to train a policy $\pi(a_t | o_t, l)$ that predicts a robot action $a_t$ given the current observation $o_t$ and a natural language instruction $l$. The observation $o_t$ consists of a set of $V$ RGB images from different camera viewpoints, denoted as $\mathbf{I_t} = \{I_1, I_2, \dots, I_V\}$, and a robot proprioceptive state $\mathbf{p}_t$. Together, these form the complete observation $o_t = (\mathbf{I_t}, \mathbf{p}_t)$. The policy is parameterized by a neural network $\pi_\theta$ with trainable weights $\theta$. The action $a_t \in \mathbb{R}^7$ at each timestep $t$ specifies the relative end-effector pose and the gripper position. This 7-dimensional action vector is defined as
$$
a_t = [\Delta x_t, \Delta y_t, \Delta z_t, \Delta \alpha_t, \Delta \beta_t, \Delta \gamma_t, g_t],
$$
where:
\begin{itemize}
    \item $(\Delta x, \Delta y, \Delta z)$ is the 3D vector for the relative translation of the end-effector.
    \item $(\Delta \alpha, \Delta \beta, \Delta \gamma)$ is the 3D axis-angle representation for the relative rotation of the end-effector.
    \item $g \in \{+1, -1\}$ is a scalar representing the gripper state, where $+1$ signifies 'open' and $-1$ signifies 'closed'.
\end{itemize}

The policy is trained on a dataset of expert demonstrations $\mathcal{D} = \{(o_t^{(i)}, l^{(i)}, a_t^{(i)})\}_{i=1}^N$ using BC, which frames the imitation learning task as a supervised learning problem. The objective is to find the optimal parameters $\theta^*$ for the policy network $\pi_\theta$ that produce actions as close as possible to the expert actions $a_t$ from the demonstration dataset $\mathcal{D}$. This is achieved by minimizing a loss function, such as the Mean Squared Error (MSE), with respect to the network parameters $\theta$:
$$
\mathcal{L}_{BC}(\theta) = \mathbb{E}_{(o_t, l, a_t) \sim \mathcal{D}} \left[ || \pi_\theta(o_t, l) - a_t ||^2 \right].
$$

Here, the expectation $\mathbb{E}$ is taken over all samples in the expert demonstration dataset $\mathcal{D}$. By minimizing this loss, the policy network learns to map observations and language commands to the corresponding expert-level actions.
  
\section{Methodology}

We propose GeoAware-VLA, a novel approach to enhance a standard VLA framework. The core of our method is the replacement of the trainable vision encoder with a frozen, Geometrically-Aware one. To make the rich, multi-layer features from the VGGT backbone usable by the policy network, we introduce a trainable vision projection layer. The remainder of the policy architecture adapts and builds upon elements from \cite{haldar2024baku}. The overall architecture of our model is illustrated in Fig.~\ref{fig:model_arch}.

The overall architecture can be divided into three stages: sensory encoding, policy decoding, and action generation. The final action $a_t$ is a function of the visual observation $o_t^{\text{vis}}$, proprioceptive state $s_t$, and language instruction $l_t$,
$$ a_t = \pi_{\phi}(P_{\theta}(E_{\text{VGGT}}(o_t^{\text{vis}})), E_{\text{lang}}(l_t), E_{\text{proprio}}(s_t)), $$
where $E$ represents the respective encoders, $P_{\theta}$ is our trainable vision projection layer, and $\pi_{\phi}$ is the policy network comprising the transformer trunk and action head.

\subsection{Sensory Encoders} 
Each input modality is processed by a specialized encoder to generate a fixed-dimension embedding, $D_{\text{repr}}$, before being passed to the policy's observation trunk.

\begin{figure}[t!]
    \centering
    \includegraphics[width=0.90\columnwidth]{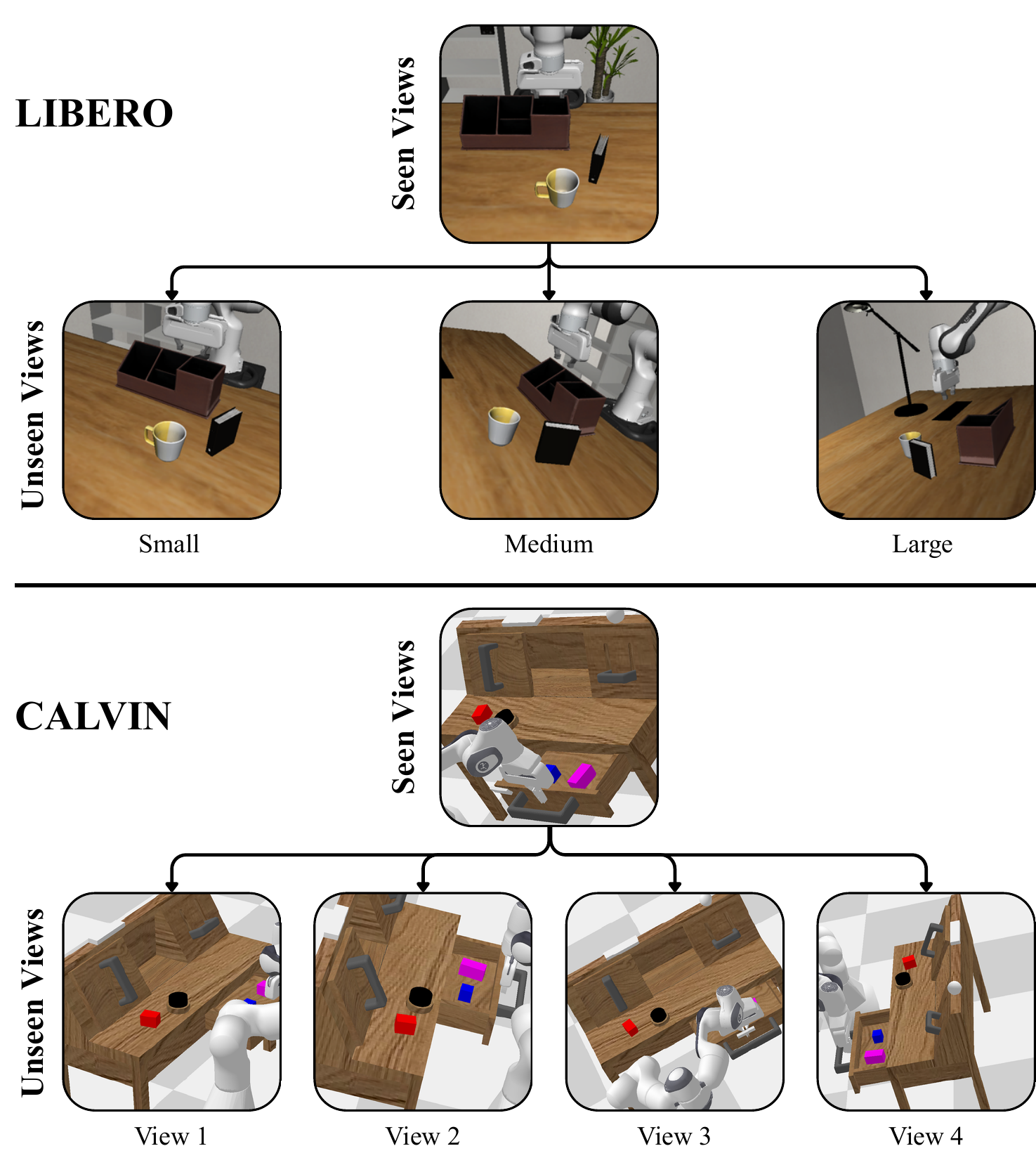}
    \caption{A visual representation of seen (used for training) and unseen views on LIBERO \textbf{(Top)} and CALVIN \textbf{(Bottom)}}
    \label{fig:unseen_views}
\end{figure}

\subsubsection{Geometrically-Aware Vision Encoder}

To encode visual observations, we utilize a frozen, pretrained VGGT model backbone and remove the prediction heads. A key property of VGGT is that it produces a list of feature tensors from multiple intermediate layers, capturing a hierarchy of visual and geometric information. Our trainable vision projection layer is designed to efficiently aggregate and condense this information.

Let the encoder output a list of $M$ feature tensors. We select a subset of evenly spaced $L$ intermediate layers for processing. For each camera view, the features from a single selected layer $l$ have a dimensionality of $N_l \times D_{\text{vggt}}$, where $N_l$ is the sequence length of visual tokens and $D_{\text{vggt}}$ is the VGGT's hidden dimension.

The vision projection layer processes features from the $L$ selected layers for each camera view. Features from each layer ($\mathbf{z}_l \in \mathbb{R}^{N_l \times D_{\text{vggt}}}$) are first passed through a dedicated 1D convolutional network with trainable convolutional and ReLU layers, followed by an adaptive average pooling layer to pool the features into a single vector ($\mathbf{f}_l \in \mathbb{R}^{D_{\text{conv}}}$). These layer-specific vectors are then concatenated into a single vector. This concatenated vector is then passed through a final multi-layer perceptron (MLP) to produce the visual embedding $\mathbf{z}_{\text{vis}} \in \mathbb{R}^{D_{\text{repr}}}$.

The overall transformation for a single view's features 
\begin{multline*}
\mathbf{z}_{\text{vis}} = \text{MLP}_{\theta} \left( \left[ \text{Conv1D}_{\theta_1}(\text{Pool}(\mathbf{z}_{l_1})); \dots; \right. \right. \\
\left. \left. \text{Conv1D}_{\theta_L}(\text{Pool}(\mathbf{z}_{l_L})) \right] \right)
\end{multline*}

This per-view projection efficiently adapts the powerful, frozen VGGT features using a lightweight, trainable module.

\subsubsection{Language and Proprioceptive Encoders}
We process non-visual modalities with simple MLP-based projectors.
\begin{itemize}
    \item \textbf{Language Encoder:} A task instruction string $l_t$ is first encoded using a pretrained sentence transformer \cite{reimers-2019-sentence-bert}, yielding an embedding of dimension $D_{\text{lang\_emb}}$. An MLP then projects this embedding into the common representation space, producing $\mathbf{z}_{\text{lang}} \in \mathbb{R}^{D_{\text{repr}}}$.
    \item \textbf{Proprioceptive Encoder:} The robot's state vector $s_t$ (end-effector pose and gripper state) is passed through a two-layer MLP to produce the
    proprioceptive embedding $\mathbf{z}_{\text{proprio}} \in \mathbb{R}^{D_{\text{repr}}}$.
\end{itemize}

\begin{figure}[t!]
    \centering
    \includegraphics[width=0.98\linewidth]{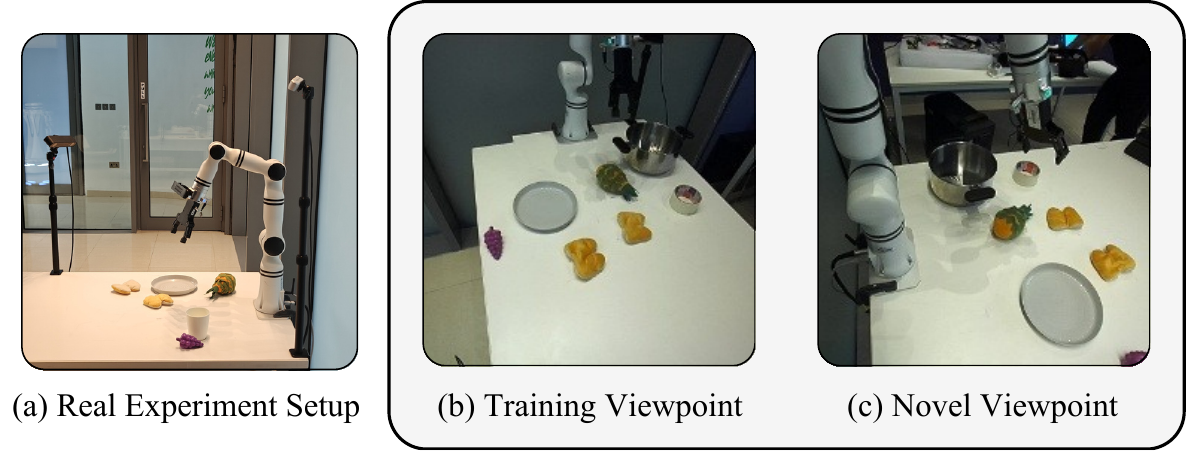}
    \caption{Real-Robot Setup. (a) Illustration of Hardware. (b) The viewpoint used to train all policies. (c) The viewpoint used for our zero-shot evaluation experiments.}
    \label{fig:real-setup}
\end{figure}

\subsection{Observation Trunk and Policy Decoder}
Our Policy decoder is a GPT-style, decoder-only transformer. The encoded representations from all modalities ($z_{\text{vis}}$ from each camera view, $z_{\text{lang}}$, and $z_{\text{proprio}}$) are treated as a sequence of input tokens. A learnable action token is appended to this sequence. The transformer processes this sequence with a causal self-attention mask, and the output embedding corresponding to the action token's position, $h_{\text{action}} \in \mathbb{R}^{D_{\text{hidden}}}$, serves as the input to the action head.

\subsection{Action Heads}
The final action is generated by a dedicated action head module conditioned on the feature vector $h_{\text{action}}$. We experiment with two variants to handle different action distributions.
\begin{itemize}
    \item \textbf{MLP Head:} A simple deterministic head, implemented as a 2-layer MLP, that directly regresses the continuous action vector $a_t \in \mathbb{R}^{D_{\text{act}}}$. This head is suitable for unimodal action distributions. we call this variant GeoAware BAKU.
    \item \textbf{VQ-BeT Head:} To potentially model multi-modal expert behavior, we use the Vector-Quantized Behavior Transformer (VQ-BeT) head. This module replaces traditional k-means clustering with a VQ-VAE to learn a discrete action codebook. It predicts an action by classifying the appropriate action token from its codebook and regressing a continuous offset, allowing it to capture complex, multi-modal action distributions. This variant is GeoAware VQ-BeT.
\end{itemize}

\begin{table*}[ht!]
\centering
\caption{Success Rates (\%) on Libero Subsets. \textbf{Bold} and \underline{underline} denote the best and second-best results, respectively. Unseen results are averaged over the three unseen views. Models marked with * are results taken from \cite{kim2025fine}.}
\label{tab:libero_results}
\begin{tabular}{lcccccccccc}
\toprule
& \multicolumn{8}{c}{Libero Subsets} & \multicolumn{2}{c}{Average} \\
\cmidrule(lr){2-9} \cmidrule(lr){10-11}
Model & \multicolumn{2}{c}{Spatial} & \multicolumn{2}{c}{Object} & \multicolumn{2}{c}{Goal} & \multicolumn{2}{c}{Long} & \multicolumn{2}{c}{Across Subsets} \\
\cmidrule(lr){2-3} \cmidrule(lr){4-5} \cmidrule(lr){6-7} \cmidrule(lr){8-9} \cmidrule(lr){10-11}
& Original & Unseen & Original & Unseen & Original & Unseen & Original & Unseen & Original & Unseen \\
\midrule
Diffusion Policy* \cite{chi2023diffusion}& 78.3 & - & 92.5 & - & 68.3 & - & 50.5 & - & 72.4 & - \\ 
Octo* \cite{ghosh2024octo} & 78.9 & - & 85.7 & - & 84.6 & - & 51.1 & - & 75.1 & - \\
OpenVLA-OFT \cite{kim2025fine} & \textbf{98.0} & \underline{90.0} & \underline{98.0} & 15.7 & \underline{98.0}  & 81.7 & 91.0 & 13.7 & \underline{96.3} & 50.2 \\

BAKU \cite{haldar2024baku} & 94.0 & 18.0 & \textbf{100.0} & 49.3 & 96.0 & 80.7 & 87.0 & 3.7 & 94.2 & 37.9 \\
VQ-BeT \cite{lee2024behavior}& 94.0 & 41.3 & \textbf{100.0} & 38.3 & \underline{98.0} & 83.0 & \textbf{93.0} & 3.0 & \underline{96.3} & 41.4 \\
Evo-0 BAKU \cite{evo0}& 77.0 & 52.0 & \textbf{100.0} & \underline{98.3} & 94.0  & \underline{88.0} & 51.0 & 28.3 & 80.5 & 66.6 \\
\midrule

GeoAware BAKU \textbf{(ours)} & \underline{95.0} & \textbf{94.3} & \underline{98.0} & 98.0 & \underline{98.0} & \textbf{90.7} & \underline{89.9} & \underline{47.3} & 95.2 & \textbf{82.6} \\
GeoAware VQ-BeT \textbf{(ours)} & \underline{95.0} & 54.3 & \textbf{100.0} & \textbf{99.0} & \textbf{99.0} & 85.7 & \textbf{93.0} & \textbf{72.7} & \textbf{96.8} & \underline{77.9} \\
\bottomrule
\end{tabular}
\end{table*}

\begin{table*}[ht!]
\centering
\caption{Success Rates (\%) on the CALVIN Benchmark Across Original and unseen Views}
\label{tab:calvin_results}
\begin{tabular}{lcccccc}
\toprule
& \multirow{2}{*}{Original View} & \multicolumn{5}{c}{Unseen Views} \\
\cmidrule(lr){3-7}
Model & & View 1 & View 2 & View 3 & View 4 & Average \\
\midrule
BAKU \cite{haldar2024baku} & 75.5 & 47.5 & 52.8 & 71.1 & 53.4 & 56.2 \\
VQ-BeT \cite{lee2024behavior}  & \underline{92.5} & 82.7 & 79.2 & \underline{91.2} & 82.0 & 83.8 \\
Evo-0 BAKU \cite{evo0} & 78.3 & 79.3 & 74.8 & 75.0 & 73.8 & 75.7 \\
\midrule
GeoAware BAKU \textbf{(ours)} & 87.8 & \underline{90.5} & \underline{93.0} & 90.8 & \underline{93.5} & \underline{91.9} \\
GeoAware VQ-BeT \textbf{(ours)} & \textbf{93.0} & \textbf{96.5} & \textbf{93.4} & \textbf{95.0} & \textbf{94.5} & \textbf{94.8} \\
\bottomrule
\end{tabular}
\end{table*}

\section{Experimental Setup}

\subsection{Simulation Benchmarks}

We evaluate our method across two simulation frameworks to assess both manipulation capabilities and visual generalization: LIBERO \cite{libero} and CALVIN \cite{mees2022calvin}. For the LIBERO benchmark, we evaluate models across all four of its task suites (Spatial, Goal, Object, and Long). Each suite consists of 10 tasks, with 50 demonstrations provided per task for training. 

Additionally, we evaluate our approach in environment D of the CALVIN simulator, featuring a 7-DoF Franka Emika Panda robot. While CALVIN provides an extensive teleoperated play dataset, we select a subset of 8 tasks that effectively capture diverse manipulation primitives: opening and closing the drawer, switching the light bulb on and off, switching the LED on and off, and moving the slider to the left and right. To maintain a consistent low-data regime across our evaluations, we restrict the training set for each CALVIN task to 50 demonstrations. 

To rigorously test visual robustness, we evaluate models trained on both benchmarks under original and unseen camera viewpoints, an approach inspired by \cite{wilcox2025adapt3r}. Specifically, we construct unseen evaluation viewpoints for all LIBERO subsets and introduce 4 distinct unseen views for the CALVIN environment. The original and unseen camera viewpoints for both simulators are illustrated in Fig.~\ref{fig:unseen_views}. All results are reported as the success rate out of 50 rollouts per task on CALVIN and on 10 per task for LIBERO.

\subsection{Real-World Experiments}

To evaluate our approach in the real world, we established a tabletop manipulation environment featuring a Realman 65B robotic arm. We collected a custom dataset by tele-operating the arm using a Meta Quest VR controller, gathering 50 demonstrations per task. The scene is captured from two fixed, distinct camera viewpoints to provide multi-view observations for the policy, results are reported over 20 rollouts per task. The setup is shown in Fig.~\ref{fig:real-setup}. We evaluate on the following five tasks:

\begin{enumerate}[label=\textbf{T\arabic*}]
    \item \textbf{``put both cups on the plate'':} Multi-object placement evaluating sequential manipulation (inspired by Libero Long).
    \item \textbf{``pick the blue cup and place it in the bowl which is on top of the plate'':} Object placement into a nested target structure.
    \item \textbf{``pick the pineapple out of the pot and place in between the cups'':} Complex spatial reasoning requiring precise placement relative to multiple objects.
    \item \textbf{``move the pineapple to the pot and the roll to the plate'':} Distinct, sequential pick-and-place operations across multiple targets (inspired by Libero Long).
    \item \textbf{``stack the blue cup on top of the white one'':} High-precision manipulation demanding stable vertical alignment, which is particularly challenging from unseen views.
\end{enumerate}
\subsection{Baselines}

We compare our \textbf{GeoAware-VLA} model against four baselines: two derived from the \cite{haldar2024baku} architecture, Evo-0 \cite{evo0} which also leverages VGGT, and OpenVLA-OFT \cite{kim2025fine} which represents a larger, pretrained model.

\begin{itemize}
    \item \textbf{BAKU:} The standard BAKU model with its original trainable vision encoder and a deterministic MLP action head.
    \item \textbf{VQ-BeT:} The BAKU model, but with its MLP head replaced by the VQ-BeT action decoder.
    \item \textbf{Evo-0 BAKU:} The BAKU model augmented with VGGT as an auxiliary spatial encoder, fusing spatial tokens with the standard vision backbone via cross-attention.
    \item \textbf{OpenVLA-OFT:} A fine-tuned variant of OpenVLA, included to evaluate against a larger, pretrained vision-language model.
\end{itemize}

\begin{figure*}[t]
    \centering    
    \includegraphics[width=\textwidth]{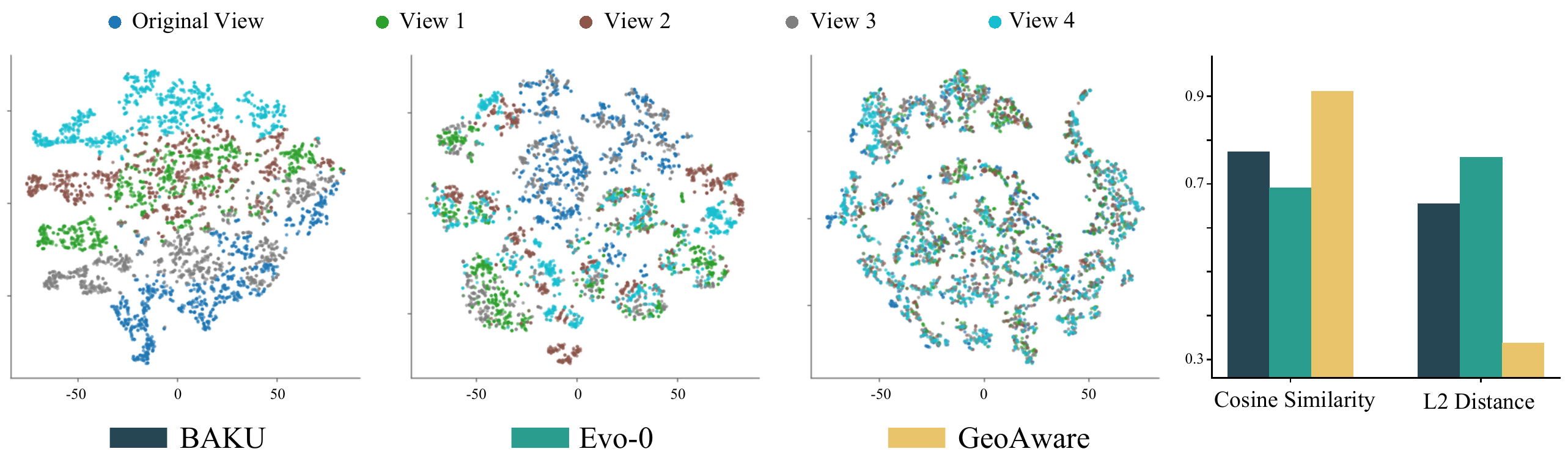}
    \caption{
    \textbf{View-invariance analysis across encoders.}
    \textbf{Left three panels:} t-SNE projections of visual embeddings for BAKU, Evo-0, and GeoAware BAKU, with points color-coded by camera view.
    Points are sampled from the same evaluation split under seen and unseen viewpoints.
    Better viewpoint invariance is indicated by stronger overlap of colors (view-agnostic structure) rather than view-separated clusters.
    \textbf{Right panel:} quantitative comparison using mean cosine similarity (\(\uparrow\), higher is better) and normalized \(L_2\) distance (\(\downarrow\), lower is better) between anchor-view and unseen-view embeddings.
    Normalized \(L_2\) is computed after unit-norm projection, making distances comparable across encoders.
    }
    \label{fig:view_invariance_analysis}
\end{figure*}

\subsection{Training}

All models were implemented in PyTorch and trained on a single NVIDIA A100 GPU with 40GB of VRAM. For experiments on the LIBERO benchmark, we trained a separate policy for each of the four task suites. Each model was trained for 150,000 steps using the AdamW optimizer. For CALVIN, all models were trained for 150 epochs. For our real-world evaluation, policies were trained for 50,000 steps on our custom dataset. Across all experiments, we used a batch size of 64 and provided the policy with observations from two camera views. For LIBERO and real-world experiments, BAKU variants were trained with input images resized to \(128\times 128\) pixels, whereas the GeoAware variants used \(126\times 126\) pixels. For CALVIN, BAKU variants used \(200\times 200\) 
pixels, and GeoAware variants used \(198\times 198\) pixels.





\begin{figure*}[t]
\centering
\begin{tikzpicture}
\begin{axis}[
    label style={font=\normalsize},
    tick label style={font=\normalsize},
    legend style={font=\normalsize, at={(0.5,1.15)}, anchor=south, legend columns=4, column sep=5pt},
    width=0.95\textwidth, 
    height=5.5cm, 
    ybar,
    ymin=0, ymax=110, 
    ytick={0,20,40,60,80,100},
    grid=both,
    major grid style={opacity=0.25},
    minor grid style={opacity=0.12},
    bar width=9pt, 
    enlarge x limits=0.1,
    symbolic x coords={T1,T2,T3,T4,T5,Avg},
    xtick=data,
    ylabel={Success Rate (\%)},
    legend cell align=left,
]

\addplot[fill=palTeal, draw=black] coordinates {
    (T1,80) (T2,100) (T3,90) (T4,70) (T5,80) (Avg,80.4)
};
\addlegendentry{BAKU (Original)}

\addplot[fill=palTeal!60, draw=black] coordinates {
    (T1,50) (T2,100) (T3,50) (T4,0) (T5,50) (Avg,50)
};
\addlegendentry{BAKU (Unseen)}

\addplot[fill=palOrange, draw=black] coordinates {
    (T1,90) (T2,100) (T3,90) (T4,90) (T5,80) (Avg,90)
};
\addlegendentry{GeoAware BAKU (Original)}

\addplot[fill=palGold, draw=black] coordinates {
    (T1,70) (T2,100) (T3,90) (T4,90) (T5,80) (Avg,86)
};
\addlegendentry{GeoAware BAKU (Unseen)}

\end{axis}
\end{tikzpicture}
\caption{Comparison of real-world performance between GeoAware-VLA and baseline. Our model demonstrates improvements across seen and unseen viewpoints.}
\label{fig:real_results}
\end{figure*}

\section{Results and Discussions}

We evaluate GeoAware-VLA to answer four key questions:
\begin{enumerate}
    \item Is leveraging a frozen, pretrained geometric encoder a viable strategy for imitation learning, even on trained-on views?
    \item How effectively does our approach generalize to unseen camera viewpoints, and how robust is this generalization to the magnitude of the viewpoint shift?
    \item Do the performance gains observed in simulation translate to physical hardware?
    \item Which layers from the geometric backbone are most critical for downstream policy performance?

\end{enumerate}

\subsection{A Geometric Backbone is a Strong In-Distribution Performer}

First, we assess whether integrating a frozen geometric backbone affects performance on the original view. As shown in Table~\ref{tab:libero_results}, our GeoAware models not only match the high success rates of their baseline counterparts on the original training views, but also yield an overall improvement in performance, boosting VQ-BeT's original performance to achieve the highest overall success rate. A similar trend can be seen on the CALVIN benchmark in Table~\ref{tab:calvin_results}. Meanwhile, other methods of integrating geometric priors like Evo-0 perform well on some of the easier subsets such as LIBERO Goal and Object but tend to lose performance on the harder task suites. This result supports our hypothesis that a lightweight, trainable projection layer is sufficient to adapt the geometrically-rich features from a frozen backbone for the policy without sacrificing in-distribution performance.

\subsection{GeoAware-VLA Achieves Robust Zero-Shot Generalization Across Viewpoint Shifts}

Our main finding is that incorporating geometric priors significantly
improves zero-shot performance on unseen camera views. As shown in
Table~\ref{tab:libero_results}, both GeoAware variants achieve gains
of over 35 percentage points on unseen viewpoints compared to their
respective baselines. Notably, the most severe performance drops for the
baseline models occur in the Spatial and Long task suites, which demand high-precision manipulation and complex spatial reasoning. A particularly informative comparison is between GeoAware BAKU and Evo-0 BAKU, as both leverage the same VGGT backbone and the same MLP action head, isolating the effect of how geometric features are integrated. While Evo-0 BAKU does improve unseen-view generalization over the standard BAKU baseline, GeoAware BAKU surpasses Evo-0 BAKU on both metrics, notably achieving 82.6\% on unseen views compared to 66.6\% for Evo-0 BAKU. This suggests that replacing the vision encoder entirely with a multi-scale geometric backbone is more effective than retaining a standard 2D encoder and fusing VGGT features as an auxiliary signal.

We observe a consistent but less pronounced trend in the CALVIN benchmark (Table~\ref{tab:calvin_results}). While the baselines still experience a performance drop under unseen camera configurations, the degradation is not as severe as in LIBERO. We attribute this to the nature of the tasks: the dominant failure mode under viewpoint shift is object mislocalization, where the policy reaches toward where an object appeared in the training view rather than where it actually is. In LIBERO, tasks such as picking up a specific object from a cluttered shelf are highly sensitive to such errors, whereas many CALVIN tasks involve coarser primitives like closing a drawer or pressing a button, where moderate localization errors are more forgivable.

To better understand the source of these gains, we analyze the learned visual representations in Fig.~\ref{fig:view_invariance_analysis}. A view-invariant encoder should produce consistent embeddings for the
same scene regardless of the observing camera pose. The t-SNE
visualizations confirm that the VGGT-based encoder produces the strongest overlap across camera views, while the baseline encoders form more view-separated clusters. We corroborate this quantitatively by computing the cosine similarity between embeddings of the same scene observed from the training view and each unseen view. GeoAware-VLA achieves a cosine similarity of 0.91, compared to 0.77 for BAKU and 0.69 for Evo-0, confirming that the geometric backbone produces a feature space that is considerably more stable across
viewpoint changes.

\subsection{Real-World Performance Gains}
We evaluate whether the performance improvements observed in simulation translate to a physical system. The results indicate that the trends transfer effectively to the real world. As detailed in Fig.~\ref{fig:real_results}, our GeoAware BAKU model achieved a measurable performance increase over the baseline on a series of manipulation tasks.

\subsection{Ablation on VGGT Layer Selection}
To determine which layers from VGGT's 24-layer backbone are most important for policy performance, we compare our default configuration (4 evenly spaced layers) with two alternatives: using all 24 layers and using only the final 4 layers.

As detailed in Table \ref{tab:vggt_ablation}, the default configuration using evenly spaced layers achieves the best overall performance. While using all layers provides a small improvement in unseen view generalization, it results in a slight drop in in-distribution performance and is more computationally expensive. It is worth noting that even the weakest configuration using the last 4 layer significantly outperforms the next-best baseline on unseen views. Despite its overall performance drop, its 34.3\% success rate on unseen views is still more than double that of \cite{kim2025fine}.
\begin{table}[t!]
\centering
\setlength{\tabcolsep}{3pt} 
\caption{Ablation study on VGGT layer selection: success rates (\%) of GeoAware VQ-BeT on the LIBERO-Long subset.}
\label{tab:vggt_ablation}
\begin{tabular}{lcc}
\toprule
Model & Original & Unseen \\
\midrule
OpenVLA-OFT \cite{kim2025fine} & \textbf{93.0} & 13.7 \\
\midrule
GeoAware VQ-BeT (All Layers) & 90.0 & \textbf{74.3} \\
GeoAware VQ-BeT (4 Evenly Spaced - Default) & \textbf{93.0} & 72.7 \\
GeoAware VQ-BeT (Last 4 Layers) & 60.0 & 34.3 \\
\bottomrule
\end{tabular}
\end{table}

\section{Conclusions and Future Work}

In this paper, we introduced GeoAware-VLA, a method that improves the generalization of VLA models to unseen camera viewpoints by replacing the standard trainable vision encoder with a pretrained geometric foundation model (VGGT) and a lightweight projection layer. This modification effectively injects a strong 3D-aware prior into the policy, relieving it from the burden of learning geometric consistency from scratch. Our experiments across the LIBERO and CALVIN benchmarks demonstrate that GeoAware-VLA matches or exceeds baseline performance on original viewpoints while achieving substantial improvements in zero-shot generalization to unseen viewpoints. These gains hold across both continuous and discrete action heads and transfer successfully to a real-world robotic platform. Our results show that the geometric capability of the vision backbone plays a critical role in building robust and generalizable robotic agents and motivates further exploration of 3D vision foundation models for imitation learning. Future work could explore alternative geometric foundation models, broader task domains and robot morphologies, and the effect of fine-tuning the geometric backbone on downstream performance.




\bibliographystyle{IEEEtran}
\bibliography{custom.bib}

\end{document}